\documentclass{article}

\PassOptionsToPackage{numbers, sort&compress}{natbib}
 \usepackage[preprint]{neurips_2026}

\usepackage[utf8]{inputenc} 
\usepackage[T1]{fontenc}    
\usepackage{url}            
\usepackage{booktabs}       
\usepackage{amsfonts}       
\usepackage{nicefrac}       
\usepackage{microtype}      

\usepackage{amsmath,amsfonts,bm}

\def\eqref#1{equation~\ref{#1}}

\def\1{\bm{1}}

\DeclareMathAlphabet{\mathsfit}{\encodingdefault}{\sfdefault}{m}{sl}
\SetMathAlphabet{\mathsfit}{bold}{\encodingdefault}{\sfdefault}{bx}{n}

\usepackage{amsmath}
\usepackage{amssymb}
\usepackage{mathtools}
\usepackage{amsthm}
\usepackage{algpseudocode}
\usepackage{algorithm}
\usepackage[table]{xcolor}

\usepackage{url}            
\usepackage{booktabs}       
\usepackage{amsfonts}       
\usepackage{nicefrac}       
\usepackage{microtype}      
\usepackage{xcolor}         
\usepackage{wrapfig}

\usepackage{microtype}
\usepackage{graphicx}
\usepackage{booktabs} 
\usepackage{multirow}
\usepackage{enumitem}
\usepackage{subcaption}
\usepackage{caption}
\usepackage{hyperref}
\hypersetup{
    colorlinks=true,
    linkcolor=black,    
    urlcolor=black,
    citecolor=black
    }

\newcommand{\ours}{ROMA}
\title{Reinforcing Multimodal Reasoning Against \\Visual Degradation}

\author{
Rui Liu$^{1,2}$,
Dian Yu$^{1}$,
Haolin Liu$^{3}$, 
Yucheng Shi$^{1}$,
Tong Zheng$^{2}$,
Runpeng Dai$^{4}$, \\ 
\textbf{Haitao Mi}$^{1}$,
\textbf{Pratap Tokekar}$^{2}$,
\textbf{Leoweiliang}$^{1}$ \\
\\ 
$^{1}$Tencent Hunyuan
$^{2}$University of Maryland, College Park \\
$^{3}$University of Virginia  
$^{4}$University of North Carolina, Chapel Hill \\
}

\begin{document}

\maketitle


\begin{abstract}

Reinforcement Learning has significantly advanced the reasoning capabilities of Multimodal Large Language Models (MLLMs), yet the resulting policies remain brittle against real-world visual degradations such as blur, compression artifacts, and low-resolution scans. Prior robustness techniques from vision and deep RL rely on static data augmentation or value-based regularization, neither of which transfers cleanly to critic-free RL fine-tuning of autoregressive MLLMs. Reinforcing reasoning against such corruptions is non-trivial: naively injecting degraded views during rollout induces reward poisoning, where perceptual occlusions trigger hallucinated trajectories and destabilize optimization. We propose ROMA, an RL fine-tuning framework that modifies the optimization dynamics to reinforce reasoning against visual degradation while preserving clean-input performance. A dual-forward-pass strategy uses teacher forcing to evaluate corrupted views against clean-image trajectories, avoiding new rollouts on degraded inputs. For distributional consistency, we apply a token-level surrogate KL penalty against the worst-case augmentation; to prevent policy collapse under regularization, an auxiliary policy gradient loss anchored to clean-image advantages preserves a reliable reward signal; and to avoid systematically incorrect invariance, correctness-conditioned regularization restricts enforcement to successful trajectories. On Qwen3-VL 4B/8B across seven multimodal reasoning benchmarks, our method improves robustness by +2.4\% on seen and +2.3\% on unseen corruptions over GRPO while matching clean accuracy.
\end{abstract}

\section{Introduction} \label{sec:intro}

Reinforcement Learning (RL) \citep{schulman2017proximal} has driven a paradigm shift in the training of large language models, unlocking strong reasoning capabilities \citep{guo2025deepseek, lambert2024t, luong2024reft, zheng2025parallel, dai2025cde, liu2026save, ouyang2022training, bai2022training}. These advances have been extended to multimodal large language models (MLLMs) \citep{li2025self, liu2025stable, liu2025vogue, huang2025vision, bai2025qwen2, bai2025qwen3}, enabling reasoning over rich visual inputs. However, such capabilities are typically developed in controlled settings with clean, well-curated data. In real-world deployment, MLLMs must contend with noisy and unstructured visual inputs, including blurry photographs, compression artifacts, and low-resolution document scans, and a model that performs reliably on a clean input (e.g., a high-quality PDF) often fails catastrophically on a degraded version of the same content. This brittleness to visual degradation poses a critical barrier to the reliable deployment of reasoning-capable MLLMs.

Visual robustness has been extensively studied in computer vision and reinforcement learning. In vision, robustness is typically pursued through data augmentation such as cropping, cutout, and flipping, often combined with contrastive objectives \citep{mumuni2024survey, sammani2023visualizing, radford2021learning}. In deep RL, a parallel line of work has shown that injecting visual augmentations during training improves out-of-distribution generalization \citep{raileanu2020automatic, yarats2021image, laskin2020reinforcement, hansen2021generalization, ma2025comprehensive}, transferring invariance learning from static perception to sequential decision-making.

Despite this progress, the visual robustness of reasoning-capable MLLMs remains underexplored, and enforcing robustness during RL fine-tuning introduces challenges that are absent in standard settings. \textbf{First, architectural mismatch.} Modern RL fine-tuning of autoregressive models increasingly relies on critic-free algorithms such as Group Relative Policy Optimization (GRPO) \citep{shao2024deepseekmath} to avoid the memory overhead of value networks; consequently, classical value-based robustness regularizers \citep{raileanu2020automatic} do not apply out of the box. \textbf{Second, reward poisoning.} Naively rolling out on degraded inputs can obscure perceptual evidence and force the model to hallucinate \citep{liu2025noisyrollout}, so the resulting reward signal penalizes perceptual failure rather than reasoning errors, destabilizing optimization and inducing policy collapse. These challenges motivate our central question: \textit{how can we make RL-fine-tuned MLLMs robust to visual degradation without sacrificing reasoning fidelity or destabilizing training?}

To answer this, we propose \textbf{ROMA}, a novel RL fine-tuning framework situated at the intersection of \textbf{M}ultimod\textbf{A}l reasoning and \textbf{RO}bust reinforcement learning. Unlike prior approaches that rely on static augmentation \citep{laskin2020reinforcement, liu2025noisyrollout, yao2025r1}, ROMA modifies the RL optimization dynamics directly to reinforce reasoning against visual degradation while preserving clean-input performance.

\begin{figure}[t]
    \centering
    \includegraphics[width=0.9\textwidth]{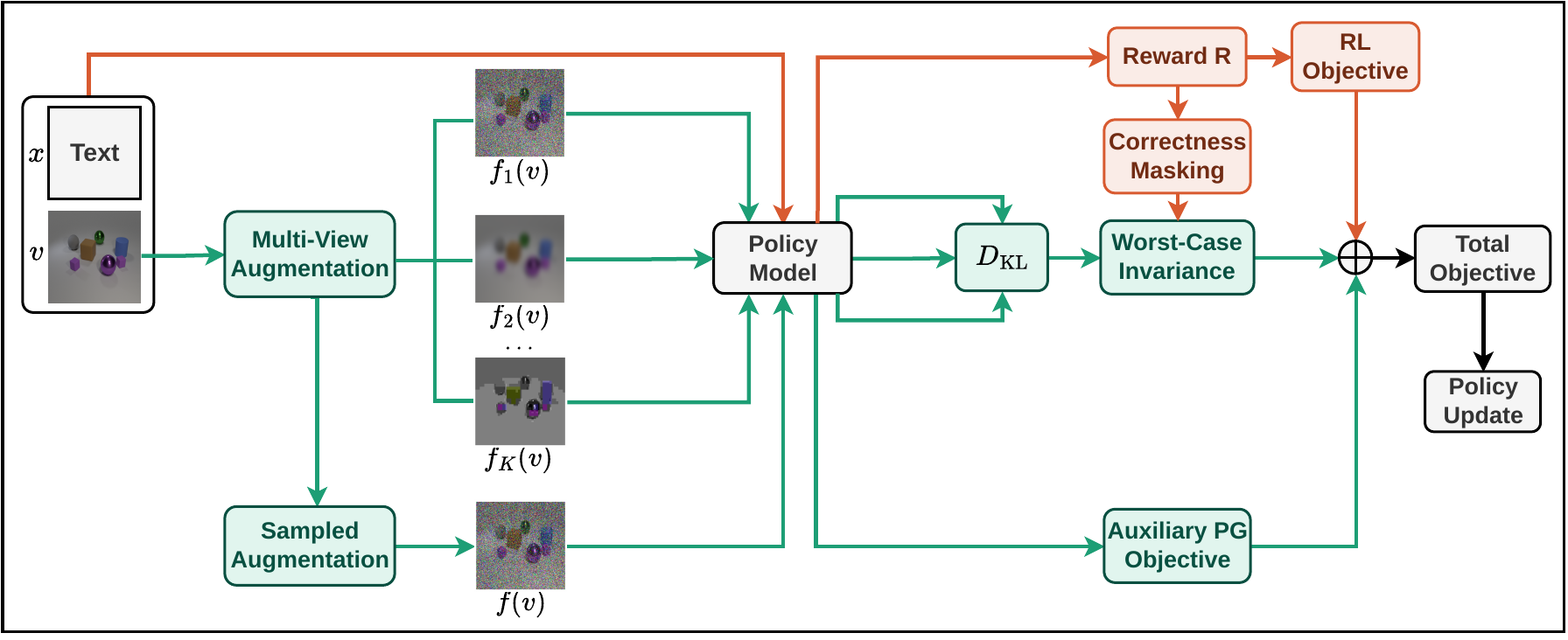}
    \caption{\textbf{Overview of ROMA}. A standard RL rollout on the clean input yields a trajectory and reward defining the main RL objective. The trajectory is then re-evaluated under perturbations via two branches: a worst-case invariance branch applying a token-level KL penalty against the most divergent of multiple degraded views, gated by a correctness mask so it fires only on successful trajectories; and an auxiliary policy gradient branch computing a clipped surrogate on a sampled degraded view, anchored to clean-image advantages. The three objectives combine into a total objective for the policy update. No rollout is sampled from a degraded input, avoiding reward poisoning.}
    \label{fig:app}
    \vspace{-10pt}
\end{figure}


At the core of our ROMA is a \emph{dual-forward-pass} training strategy over a critic-free autoregressive MLLM, as illustrated in Figure \ref{fig:app}. The first pass performs standard RL rollouts on the clean image, producing reasoning trajectories and their advantages. The second pass generates multiple degraded views of the same image and re-evaluates the \emph{same} frozen trajectory via teacher forcing, computing token-level log-probabilities under each corrupted view without sampling new rollouts. This sidesteps reward poisoning by construction: trajectories are never sampled from degraded inputs, yet we still observe how the model's token distributions shift under perturbation.

On top of this scaffold, ROMA introduces three regularizers that together yield robust reasoning. (i) A token-level surrogate KL penalty enforces distributional consistency between clean and degraded views, applied in a \emph{worst-case} fashion against the augmentation with the largest divergence. (ii) An auxiliary policy gradient loss is computed on a randomly sampled degraded view but \emph{anchored to clean-image advantages}, preserving a reliable reward signal and preventing collapse under regularization. (iii) \emph{Correctness-conditioned} regularization restricts invariance enforcement to successful trajectories, so the model is not pushed toward becoming consistently but systematically incorrect.

We validate ROMA by fine-tuning Qwen3-VL 4B and 8B Instruct models \citep{bai2025qwen3} and evaluating visual robustness across seven multimodal reasoning benchmarks: MathVista \citep{lu2023mathvista}, WeMath \citep{qiao2024we}, ChartQA \citep{masry2022chartqa}, LogicVista \citep{xiao2024logicvista}, MMStar \citep{chen2024rightwayevaluatinglarge}, VisualPuzzles \citep{song2025visualpuzzles}, and RealWorldQA \citep{grok15v}. While standard GRPO reaches strong clean-input accuracy (68.9\% at 8B), it degrades sharply under corruption, falling to 59.2\% on seen and 54.0\% on unseen perturbations. ROMA matches clean performance (68.7\%) while substantially improving robustness, reaching 61.6\% on seen (+2.4\%) and 56.3\% on unseen (+2.3\%) perturbations, with consistently smaller clean-to-degraded gaps. 

In summary, our key contributions are as follows:
\begin{itemize}[left=0pt]
\vspace{-5pt}
\item We propose \ours, an RL fine-tuning approach for MLLMs that enforces robustness to visual degradation. 
\item \ours~combines a correctness-conditioned, token-level KL invariance penalty applied in a worst-case multi-view manner with an auxiliary policy gradient anchored to clean advantages, enabling stable robustness learning in critic-free settings.
\item \ours~improves robustness on seven multimodal benchmarks empirically, achieving higher accuracy under both seen and unseen corruptions while maintaining strong clean-input performance.
\end{itemize}

\vspace{-8pt}
\section{Related Work} \label{sec:related_work}

\vspace{-6pt}
\paragraph{Visual Robustness and Data Augmentation in RL.}

The pursuit of visual robustness via data augmentation has a long history in deep reinforcement learning . Methods such as Data-regularized Actor-Critic (DrAC) \citep{raileanu2020automatic}, RAD \citep{laskin2020reinforcement}, and DrQ \citep{yarats2021image} demonstrate that applying visual augmentations, such as cropping, blurring, or flipping, can improve OOD generalization. In these traditional actor-critic setups, robustness is achieved by regularizing both the policy and the value networks to maintain consistent representations across clean and augmented states, allowing agents to generalize effectively to novel environments \citep{hansen2021generalization, ma2025comprehensive}.

Despite their success in continuous control and standard discrete environments, these traditional regularization techniques are fundamentally incompatible with modern MLLM fine-tuning due to architectural mismatches and the semantic sensitivity of multimodal reasoning. Our work advances the paradigm by reformulating visual invariance specifically for large-scale, critic-free generative models. Instead of relying on a value network, we introduce a token-level surrogate KL divergence penalty. Moreover, rather than applying uniform augmentation, we employ a worst-case multi-view strategy that focuses optimization on the most adversarial corruption at each step. Combined with an auxiliary policy gradient objective, our approach enables robust invariances learning while preserving the semantic and logical consistency required for multimodal reasoning.

\vspace{-6pt}
\paragraph{Reinforcement Learning for Multimodal Reasoning.}

Reinforcement learning has recently emerged as a powerful paradigm for eliciting complex reasoning in MLLMs. For instance, \citet{tan2025reason} adapt text-based reasoning paradigms to multimodal settings, while \citet{peng2025lmm} scale mathematical reasoning and cross-modality alignment. Concurrently, \citet{yang2025r1} extend language paradigms to improve visual question answering, and \citet{huang2025vision} employ vision-grounded prompts to facilitate multi-step logic. More recently, a line of research has begun to investigate MLLMs reasoning leveraging visual perturbations. To ensure models rely on visual context rather than linguistic priors, \citet{wang2025perception} encourage visual grounding by penalizing the policy when its outputs remain unchanged under heavy masking. \citet{liu2025noisyrollout} attempt to reinforce visual exploration by directly injecting data augmentation into the environment during the RL generation phase. Furthermore, \citet{liu2025vogue} utilize visual uncertainty to guide policy exploration.

Despite these advancements, robustness to visual degradation in RL-based multimodal reasoning remains underexplored. Our approach addresses this gap by explicitly targeting both robustness and OOD generalization in MLLM reasoning. We introduce a correctness-conditioned, token-level invariance penalty tailored for critic-free frameworks, ensuring that reasoning trajectories remain resilient to visual noise. Moreover, unlike standard RL fine-tuning, which can inadvertently reinforce hallucinated reasoning under perceptual occlusion, our approach anchors the advantage computation to clean visual states. This prevents the reward poisoning common in naive data augmentation, preserving the logical integrity of the learned policy.

\vspace{-6pt}
\section{Approach} \label{sec:approach}
\vspace{-6pt}
In this section, we present our approach for improving the visual robustness and OOD generalization of MLLMs trained via RL. We first formalize the autoregressive fine-tuning setting, and then introduce our key components: a correctness-conditioned, token-level invariance regularization objective, and a worst-case multi-view optimization strategy combined with an auxiliary policy gradient objective to enforce robustness.

\paragraph{Problem Formulation.}
\label{subsec:problem_formulation}

We consider a multimodal reasoning task where a MLLM produces a logical chain-of-thought to answer a visual query. Each input consists of a text question $x$ and an associated image $v$. To solve the task, the MLLM acts as a stochastic policy $\pi_\theta$, parameterized by $\theta$, generating a step-by-step reasoning trajectory $y \sim \pi_\theta(\cdot \mid v, x)$. Upon generating the complete trajectory $y$, a reward function evaluates its correctness and yields a scalar reward $R(v, x, y)$. The standard reinforcement learning objective seeks to maximize this expected reward:
\begin{equation}
    J_{\text{RL}}(\theta) = \mathbb{E}_{(v, x) \sim \mathcal{D}, \, y \sim \pi_\theta(\cdot \mid v, x)} \left[ R(v, x, y) \right]
\end{equation}


However, optimizing this objective solely on clean images $v$ leads to policies that fail to generalize under real-world visual degradations (e.g., blur, sensor noise, and compression artifacts). Consequently, our goal is to regularize $\pi_\theta$ such that the generated trajectory $y$ remains robust and logically consistent even under degraded visual inputs.

\paragraph{Correctness-Conditioned Token-Level Invariance.}
\label{subsec:token_level_drac}

To embed visual invariance directly into the autoregressive generation process, we draw inspiration from \citep{raileanu2020automatic}. Traditional actor-critic methods enforce invariance jointly across both policy and value networks. However, modern large-scale RL frameworks (e.g., GRPO \citep{shao2024deepseekmath}) are inherently critic-free, making value-based regularization inapplicable. We therefore isolate the policy invariance objective and reformulate it as a token-level surrogate KL divergence penalty tailored to autoregressive generation.

Let $f \in \mathcal{F}$ be a stochastic visual augmentation function, such that $f(v)$ produces a degraded view of the original input $v$. To enforce perceptual invariance, the token distribution under the degraded view should align with that of the clean view. Treating the clean visual state as a reference anchor, we penalize the divergence between the degraded and clean policy logits. To prevent the noisy gradients from corrupting the clean representations, we apply a stop-gradient operator ($\text{sg}[\cdot]$) to the clean policy outputs. For a given trajectory $y$ sampled from the old policy $\pi_{\text{old}}$, the invariance penalty is defined as:
\begin{equation}
    G_\pi(\theta, f) = \mathbb{E}_{(v,x) \sim \mathcal{D}, \, y \sim \pi_{\text{old}}} \left[ \sum_{t=1}^{|y|} 
    D_{\text{KL}} \Big( \text{sg} \big[ \pi_\theta(\cdot \mid v, x, y_{<t}) \big] \,\|\, \pi_\theta(\cdot \mid f(v), x, y_{<t}) \Big)
    \right],
\end{equation}
where the per-token KL divergence is practically approximated via the standard RL surrogate:
$
    D_{\text{KL}}^{(t)} \approx p_t \cdot (\log p_t - \log q_t),
$
with $p_t = \text{sg} \big[ \pi_\theta(y_t \mid v, x, y_{<t}) \big]$ and $q_t = \pi_\theta(y_t \mid f(v), x, y_{<t})$. Crucially, enforcing consistency across views is actively harmful if the underlying trajectory $y$ is hallucinated or factually incorrect. To prevent the policy from becoming robustly incorrect, we introduce a correctness mask, applying the penalty strictly to trajectories that successfully solve the task ($R > 0$).

\paragraph{Worst-Case Multi-View Optimization.}
\label{subsec:worst_case_multi_view}

During standard training, randomly sampled augmentations may be visually trivial, providing weak regularization signals. To enforce rigorous adversarial robustness, we depart from single-view augmentation in favor of a worst-case multi-view strategy.
      
At each training step, we sample a subset of $K$ distinct augmentations, $\mathcal{F}_K \subset \mathcal{F}$, generating $K$ degraded views. We compute the token-level invariance penalty $G_\pi(\theta, f_k)$ for all $K$ views. Rather than averaging these penalties, we apply a minimax formulation, regularizing the policy exclusively against the augmentation that induces the maximum divergence:
\begin{equation} \label{eq:worst}
     G_\pi^{\text{worst}}(\theta) = \max_{f_k \in \mathcal{F}_K} G_\pi(\theta, f_k).
\end{equation}

\paragraph{ Auxiliary Policy Gradient Loss.} \label{subsec: aux_pg}

While $G_\pi^{\text{worst}}$ enforces distributional consistency, excessive KL regularization without a grounding reward signal can induce policy collapse, where the MLLM learns to output consistent but nonsensical tokens. To provide an active learning  signal under degradation, we introduce an auxiliary policy gradient objective ($J_{\text{aug\_pg}}$). We compute an additional clipped-surrogate objective directly on the augmented logits of a randomly sampled view. Crucially, to prevent reward poisoning, we evaluate this objective using the exact token trajectories $y$ and advantages $A(v, x, y)$ derived from the clean rollout:
\begin{equation}
    J_{\text{aug\_pg}}(\theta) = \mathbb{E}_{f \sim \mathcal{F}_K} \left[ \mathbb{E} \left[ \sum_{t=1}^{|y|} \min \left( \rho_t A, \text{clip}(\rho_t, 1-\epsilon, 1+\epsilon) A \right) \right] \right],
\end{equation}
where $f \sim \mathcal{F}_K$ is a randomly sampled augmentation function from the augmentation pool, and the importance sampling ratio is $\rho_t = \frac{\pi_\theta(y_t \mid f(v), x, y_{<t})}{\pi_{\text{old}}(y_t \mid v, x, y_{<t})}$. By anchoring both the rollout generation and the advantage computation to the clean images, we force the model to actively maximize the expected reward under visual noise without training on structurally hallucinated exploration paths.

The final consolidated optimization objective for our robustness training is formulated as follows:
\begin{equation}
       J_{\text{total}}(\theta) = J_{\text{RL}}(\theta) + \alpha \cdot J_{\text{aug\_pg}}(\theta) - \beta 
      \cdot \mathbb{E} \left[ G_\pi^{\text{worst}}(\theta) \cdot \mathbb{I}[R(v, x, y) > 0] \right]
\end{equation}
where $J_{\text{RL}}(\theta)$ represents the main reinforcement learning objective (e.g., GRPO), $\alpha$ and $\beta$ are coefficients controlling the strength of the worst-case invariance penalty and auxiliary optimization, respectively. Ultimately, we update the policy parameters $\theta$  to maximize $J_{\text{total}}(\theta)$. This unified objective simultaneously drives the MLLM to maximize logical reasoning performance on clean inputs ($J_{\text{RL}}$), actively learn robust feature representations under visual degradation ($J_{\text{aug\_pg}}$), and minimize the worst-case distributional divergence between the clean and degraded reasoning paths ($-G_\pi^{\text{worst}}$).

\vspace{-6pt}
\section{Experiments}
\label{sec:experiments}
\vspace{-6pt}

To evaluate the effectiveness of our proposed framework, we design experiments to answer the following questions: (1) Does our approach improve the robustness of MLLMs against visual degradation? (2) Does the framework generalize to out-of-distribution (OOD) visual corruptions not seen during training? (3) How do individual components, such as worst-case optimization, auxiliary policy gradients, and correctness-conditioning, contribute to the overall performance?

\vspace{-6pt}
\subsection{Experimental Setup} \label{subsec:experimental_setup}

\vspace{-6pt}
\paragraph{Implementation Details.} \label{subsec:implementation_details}
We conduct direct RL training on the Qwen3-VL-4B and 8B Instruct \citep{bai2025qwen3} models, using GRPO as the underlying RL algorithm. The models are trained to generate responses in a structured format, where the reasoning process is enclosed within \(\texttt{<thinking></thinking>}\) tags and the final answer is presented in \texttt{\textbackslash boxed\{\}}. For our robustness framework, we set the multi-view sample size to $K=3$ augmentations per step. The auxiliary augmented policy gradient coefficient $\alpha$ is set to $0.10$, and the worst-case invariance regularization coefficient $\beta$ is set to $0.10$. Please see  a series of sensitivity analysis for these values in Section \ref{subsec:sensitivity_analysis}. The implementation is built on the EasyR1 framework \citep{zheng2025easyr1}. More implementation details can be found in Appendix \ref{app:imp}. 

\vspace{-6pt}
\paragraph{Dataset and Evaluation.}  We train all models on the MMRL30k dataset \citep{zhu2025shuffle}, which contains around 30K samples. We evaluate on seven multimodal reasoning benchmarks, including MathVista \citep{lu2023mathvista}, WeMath \citep{qiao2024we}, ChartQA \citep{masry2022chartqa}, LogicVista \citep{xiao2024logicvista}, MMStar \citep{chen2024rightwayevaluatinglarge}, VisualPuzzles \citep{song2025visualpuzzles}, and RealWorldQA \citep{grok15v}. These benchmarks cover a diverse range of multimodal reasoning, including mathematical problem solving, chart understanding, general visual reasoning, and logical inference. For evaluation, we use Qwen2.5-72B-Instruct \citep{qwen2.5} to extract final answers from model responses and assess their correctness against reference answers following prior work~\citep{zhu2025shuffle, liu2025vogue, liu2025stable}. 

\vspace{-6pt}
\paragraph{Baselines.} We evaluate our approach against two controlled baselines: (1) Base model: the pre-trained, instruction-tuned model prior to any RL fine-tuning. (2) GRPO: a model fine-tuned via standard GRPO on clean data. In addition, for broader context, we include evaluated results from several external models, including NoisyRollout-7B \citep{liu2025noisyrollout}, PAPO-7B \citep{wang2025perception}, Vision-R1-7B \citep{huang2025vision}, VL-Rethinker-7B \citep{wang2025vl}, and OpenVLThinker-7B \citep{deng2025openvlthinker}. Vision-R1-7B used WeMath as training data, its performance on that benchmark is omitted.

\vspace{-6pt}
\paragraph{Degradation Protocols.} We systematically evaluate our approach across three settings: (1) Clean, (2) Seen degradations, and (3) Unseen degradations. The seen setting addresses Question 1 by measuring robustness against the types of visual degradations experienced during training. Inspired by the ImageNet-C framework \citep{hendrycks2019benchmarking}, this pool simulates common image capture and transmission artifacts: Gaussian noise, Gaussian blur, JPEG compression, and resolution downscaling. Conversely, the unseen setting addresses Question 2 by assessing OOD generalization across novel corruption types. This pool subjects the model to corruptions strictly held out during training: motion blur, salt-and-pepper noise, speckle noise, posterization, and pixelation. Detailed degradation parameters and visual examples are provided in Appendix \ref{app:deg} and Figure \ref{fig:deg}. Crucially, for the main results, we evaluate performance at a severe magnitude (Level 3) that strictly exceeds the parameter bounds used during training, thereby testing the model's ability to extrapolate to unseen severity distributions.
      




\begin{table*}[t]
    \centering
    \caption{\textbf{Robustness performance evaluation with the 4B model}. To present a consolidated view of visual robustness, we report performance on degraded inputs as macro-averages across both seen and unseen degradation types. Our approach demonstrates superior robustness, achieving the highest average accuracy across both seen and unseen degradations.} 
    \vspace{-5pt}
    \label{tab:main_results_4b}
    \resizebox{\textwidth}{!}{
    \begin{tabular}{l ccccccc c}
        \toprule
        \textbf{Method} & \textbf{MathVista} & \textbf{WeMath} &  \textbf{ChartQA} & \textbf{LogicVista} & \textbf{MMStar} & \textbf{VisPuzzles} & \textbf{RealWorldQA} & \textbf{Avg} \\
        \midrule
        \multicolumn{9}{c}{\textit{Clean}} \\
        \midrule
        Qwen3-VL-4B Instruct \citep{bai2025qwen3} & 76.6 & 64.7 & 79.2 & 59.6 & 66.3 & 41.9 & 69.1 & 65.3 \\
        \quad + GRPO      & 78.0 & 76.6 & 81.3 & 58.0 & 69.1 & 40.8 & 70.1 & 67.7 \\
        \quad + \ours & 78.4 & 76.4 & 80.6 & 60.0 & 69.9 & 42.3 & 69.5 & 68.2 \\
        \midrule
        \multicolumn{9}{c}{\textit{Seen Degradations}} \\
        \midrule
        Qwen3-VL-4B Instruct \citep{bai2025qwen3} & 70.3 & 63.9 & 46.3 & 52.6 & 64.6 & 39.9 & 65.2 & 57.5 \\
        \quad + GRPO       & 70.7 & 74.3 & 48.0 & 52.4 & 64.0 & 39.2 & 64.6 & 59.0 \\
        \quad + \textbf{\ours} & \textbf{73.1} & \textbf{75.0} & \textbf{48.4} & \textbf{55.6} & \textbf{66.0} & \textbf{41.5} & \textbf{65.5} & \textbf{60.7} \\
        \midrule
        \multicolumn{9}{c}{\textit{Unseen Degradations}} \\
        \midrule
        Qwen3-VL-4B Instruct \citep{bai2025qwen3} & 61.8 & 55.7  & 41.2 & 48.5 & 59.1 & 35.3 & 60.6 & 51.7 \\
        \quad + GRPO       & 62.9 & 65.6 & 42.4 & 48.5 & 59.6 & 36.6 & 61.2  & 53.8 \\
        \quad + \textbf{\ours} & \textbf{64.5} & \textbf{67.0} & \textbf{43.4} & \textbf{49.6} & \textbf{61.1} & \textbf{37.2} & \textbf{62.6} & \textbf{55.1} \\
        \bottomrule
    \end{tabular}
    }
    \vspace{-10pt}
\end{table*}

\vspace{-6pt}
\subsection{Main Results} \label{subsec:main_results}

Tables \ref{tab:main_results_4b} and \ref{tab:main_results_8b} present the main evaluation results for the Qwen3-VL 4B and 8B Instruct models, respectively. To provide a consolidated view of visual robustness, results under degradation are reported as macro-averages across all specific perturbation types within the seen and unseen pools for each dataset. For a detailed breakdown of performance under each specific degradation type, please refer to Appendix \ref{app:exp}. 

We first establish the baseline performance on clean data. As shown, standard GRPO yields solid improvements over the base model on clean data, achieving an average score of 67.7\% (compared to the 4B base model's 65.3\%) and 68.9\% (compared to the 8B base model's 66.8\%). Our approach performs comparably to GRPO on these clean inputs for both the 4B (68.2\%) and 8B (68.7\%) models. This demonstrates that our anchored optimization framework successfully preserves foundational reasoning capabilities without compromising baseline performance.

\vspace{-6pt}
\paragraph{Robustness to Visual Degradations.}
We next evaluate the models under the Seen degradation setting to measure visual robustness. As detailed in the Tables \ref{tab:main_results_4b} and \ref{tab:main_results_8b}, GRPO suffers a larger performance drop when transitioning from clean to degraded inputs, decreasing by 8.7\% (from 67.7\% to 59.0\%) for the 4B model, and by 9.7\% (from 68.9\% to 59.2\%) for the 8B model. Standard GRPO struggles to maintain performance under visual perturbations. In contrast, our approach consistently outperforms GRPO across all benchmarks under degraded conditions. The performance gap between clean and degraded inputs for our 8B model is reduced to a drop of 7.1\%, compared to the 9.7\% drop observed in GRPO. By anchoring the advantage computation to clean inputs and penalizing structural deviation via the token-level invariance penalty, our framework successfully mitigates the impact of perceptual artifacts encountered during training.

\begin{table*}[ht]
    \centering
    \caption{\textbf{Robustness performance evaluation with the 8B model}. To present a consolidated view of visual robustness, we report performance on degraded inputs as macro-averages across both seen and unseen degradation types. Our approach demonstrates superior robustness, achieving the highest average accuracy across both seen and unseen degradations.} 
    \vspace{-5pt}
    \label{tab:main_results_8b}
    \resizebox{\textwidth}{!}{
    \begin{tabular}{l ccccccc c}
        \toprule
        \textbf{Method} & \textbf{MathVista} & \textbf{WeMath} & \textbf{ChartQA} & \textbf{LogicVista} & \textbf{MMStar} & \textbf{VisPuzzles} & \textbf{RealWorldQA} & \textbf{Avg} \\
        \midrule
        \multicolumn{9}{c}{\textit{Clean}} \\
        \midrule
        
        OpenVLThinker-7B \citep{deng2025openvlthinker} & 67.0 & 60.6 & 78.7 & 48.0 & 60.1 & 32.0 & 58.6 & 57.9 \\
        Vision-R1-7B \citep{huang2025vision} & 72.4 & -- & 81.6 & 48.7 & 62.7 & 36.1 & 66.1 & -- \\
        NoisyRollout-7B \citep{liu2025noisyrollout} & 72.7 & 69.3 & 79.8 & 50.0 & 63.2 & 37.7 & 67.1 & 62.8 \\
        PAPO-7B \citep{wang2025perception} & 75.5 & 71.0 & 82.0 & 53.3 & 63.2 & 35.8 & 67.3 & 64.0 \\
        VL-Rethinker-7B \citep{wang2025vl} & 72.7 & 67.5 & 79.9 & 46.9 & 61.9 & 34.8 & 68.5 & 61.7 \\
        
        Qwen3-VL-8B Instruct \citep{bai2025qwen3} & 76.6 & 69.4 & 79.4 & 60.7 & 68.7 & 43.5 & 69.4 & 66.8 \\
        \quad + GRPO       & 78.4 & 77.6 & 81.5 & 60.8 & 70.1 & 43.5 & 70.6 & 68.9 \\
        \quad + \ours & 78.5 & 77.9 & 80.8 & 62.1 & 69.5 & 42.5 & 69.9 & 68.7 \\
        \midrule
        \multicolumn{9}{c}{\textit{Seen Degradations}} \\
        \midrule
        OpenVLThinker-7B \citep{deng2025openvlthinker} & 60.6 & 60.5 & 45.1 & 44.4 & 56.3 & 32.8 & 54.8 & 50.6 \\
        Vision-R1-7B \citep{huang2025vision} & 63.9 & -- & 47.4 & 46.0 & 58.9 & 34.8 & 60.6 & -- \\
        NoisyRollout-7B \citep{liu2025noisyrollout} & 66.9 & 67.4 & 46.1 & 46.5 & 60.4 & 34.6 & 62.1 & 54.9 \\
        PAPO-7B \citep{wang2025perception} & 68.2 & 70.0 & 47.3 & 47.2 & 59.8 & 34.2 & 60.9 & 55.4 \\
        VL-Rethinker-7B \citep{wang2025vl} & 64.4 & 67.1 & 45.9 & 46.0 & 58.1 & 33.8 & 63.1 & 54.1 \\
        
        Qwen3-VL-8B Instruct \citep{bai2025qwen3} & 70.7 & 69.4 & 46.9 & 54.6 & 64.8 & 40.1 & 65.7 & 58.9  \\
        \quad + GRPO       & 71.2 & 75.0 & 48.0 & 52.7 & 64.0 & 38.9 & 64.8 & 59.2 \\
        \quad + \textbf{\ours} & \textbf{73.3} & \textbf{77.3} & \textbf{49.1} & \textbf{57.5} & \textbf{66.3} & \textbf{41.7} & \textbf{66.0} & \textbf{61.6} \\
        \midrule
        \multicolumn{9}{c}{\textit{Unseen Degradations}} \\
        \midrule

        OpenVLThinker-7B \citep{deng2025openvlthinker} & 54.4 & 54.7 & 41.6 & 41.8 & 51.6 & 30.1 & 50.9 & 46.4 \\
        Vision-R1-7B \citep{huang2025vision} & 58.3 & -- & 44.8 & 42.0 & 54.3 & 31.6 & 57.1 & -- \\
        NoisyRollout-7B \citep{liu2025noisyrollout} & 60.5 & 62.1 & 43.0 & 42.1 & 55.4 & 30.8 & 56.7 & 50.1 \\
        PAPO-7B \citep{wang2025perception} & 60.8 & 63.6 & 42.6 & 43.3 & 55.5 & 31.4 & 57.4 & 50.7 \\
        VL-Rethinker-7B \citep{wang2025vl} & 57.6 & 61.6 & 43.1 & 41.8 & 53.6 & 30.2 & 59.0 & 49.6 \\
        
        Qwen3-VL-8B Instruct \citep{bai2025qwen3} & 63.1 & 60.4 & 42.1 & 50.0 & 59.5 & 37.3 & 61.4 & 53.4 \\
        \quad + GRPO       & 63.0 & 66.8  & 43.0 & 47.1 & 59.8 & 36.8 & 61.4 & 54.0 \\
        \quad + \textbf{\ours} & \textbf{64.8} & \textbf{70.4} & \textbf{44.1} & \textbf{50.8} & \textbf{62.1} & \textbf{38.0} & \textbf{63.6} & \textbf{56.3} \\
        \bottomrule
    \end{tabular}
    }
    \vspace{-10pt}
\end{table*}

\paragraph{Generalization to OOD Degradations.}
Furthermore, we evaluate the OOD generalization of our approach on degradation types completely unseen during training. As shown in Table \ref{tab:main_results_4b} and Table \ref{tab:main_results_8b}, our approach exhibits stronger zero-shot generalization to these unseen corruptions for both model sizes. Under OOD conditions, the 8B GRPO performance drops to 54.0\%. However, our framework sustains an average score of 56.3\%, outperforming the standard RL baseline. Additionally, the performance drop from clean to OOD evaluation is 12.4\% for our 8B method, which is smaller than the 14.9\% decrease observed in GRPO. This confirms that the robustness acquired on seen degradations transfers effectively to unseen domains, validating that our token-level constraint encourages generalized resilience without overfitting to the training distribution.

\begin{figure*}[ht]
    \centering
    \begin{subfigure}[t]{0.42\linewidth} 
        \centering
        \includegraphics[width=\textwidth]{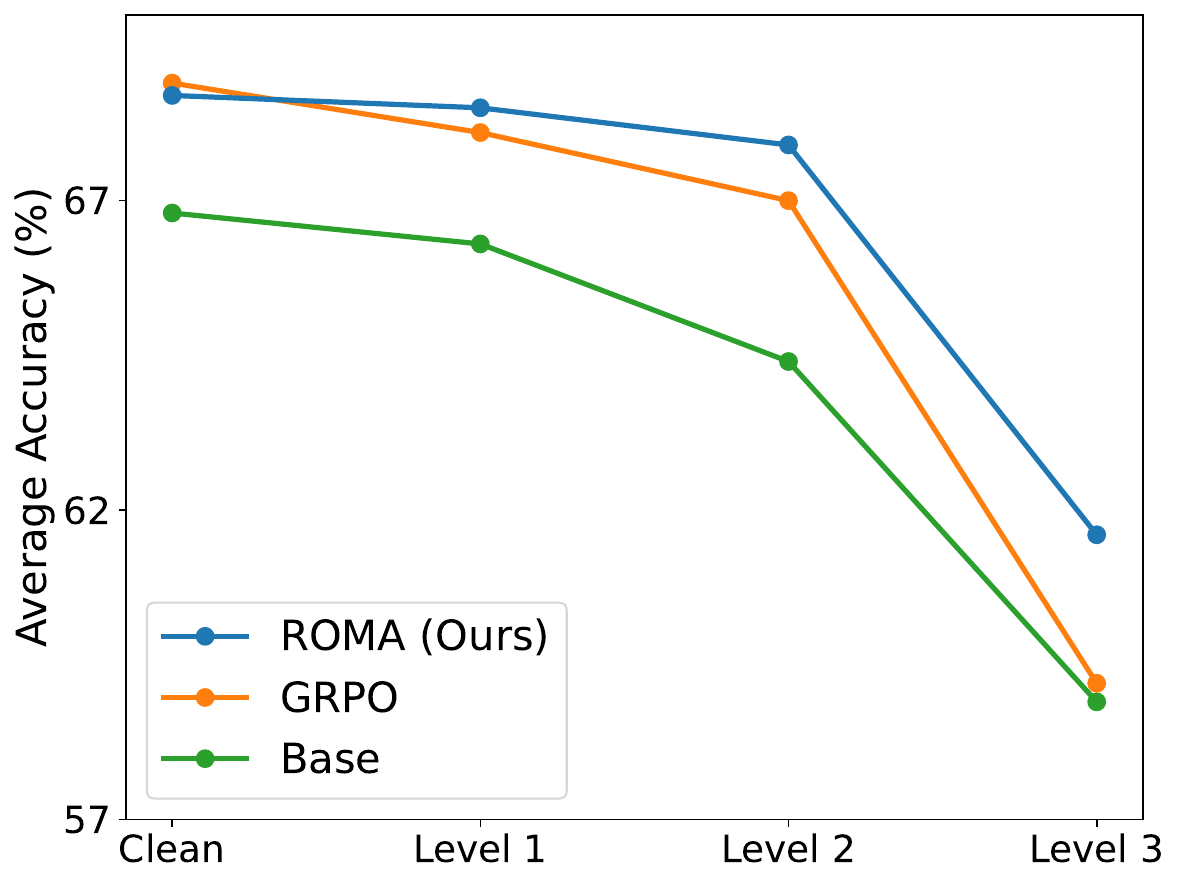}
        \caption{Evaluation on seen degradations.}
        \label{fig:verifer_acc}
    \end{subfigure}
    \begin{subfigure}[t]{0.42\linewidth}
        \centering
        \includegraphics[width=\textwidth]{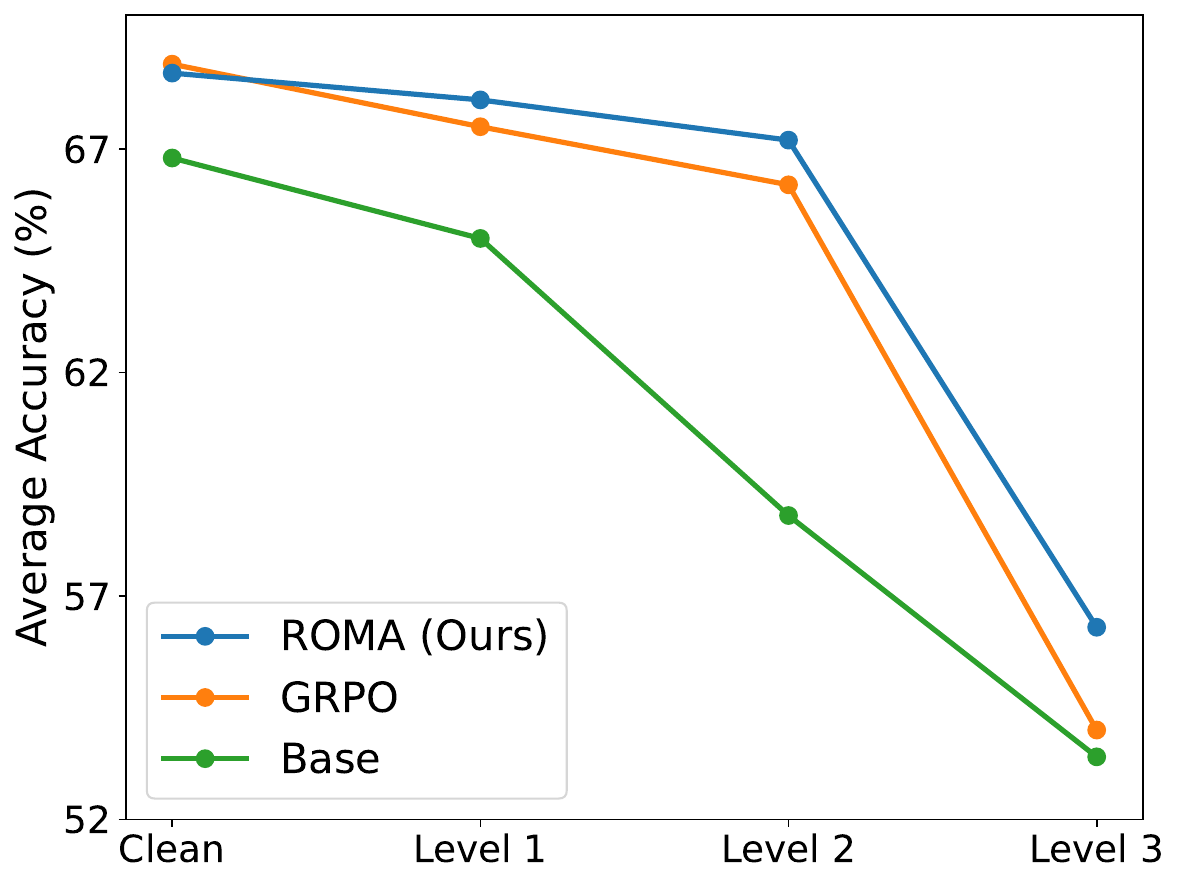}
        \caption{Evaluation on unseen degradations.}
        \label{fig:val_acc}
    \end{subfigure}
    \caption{\textbf{Robustness under increasing visual corruption severity}. We report accuracy from Clean to Level 3 (severe) for seen and unseen degradations. \ours~consistently achieves higher accuracy at severe corruption levels and exhibits smaller performance degradation compared to the base model and GRPO.}
    \label{fig:level}
    \vspace{-8pt}
\end{figure*}

\paragraph{Performance Across Degradation Levels.}
We further evaluate robustness by measuring accuracy under progressively stronger visual corruptions, from Clean to Level 3 (severe), as illustrated in Figure~\ref{fig:level}. On seen degradations, the 8B base model drops from 66.8\% to 58.9\% (-7.9\%), while GRPO declines from 68.9\% to 59.2\% (-9.7\%). In contrast, our method decreases from 68.7\% to 61.6\% (-7.1\%), achieving the highest accuracy at Level 3 and outperforming GRPO by +2.4\%. On unseen degradations, the base model exhibits a larger degradation from 66.8\% to 53.4\% (-13.4\%), and GRPO drops from 68.9\% to 54.0\% (-14.9\%). Our method again demonstrates superior robustness, decreasing from 68.7\% to 56.3\% (-12.4\%), outperforming GRPO by +2.3\% at the most severe corruption Level 3. 

Overall, while all methods degrade under increasing corruption, our approach demonstrates smaller performance drops and stronger final accuracy, indicating improved robustness to both seen and unseen visual perturbations.

\vspace{-6pt}
\subsection{Ablation Studies} \label{subsec:ablation_studies}
To address Question 3 and validate our design configurations, we conduct a series of ablation studies. Specifically, we analyze the choice of multi-view optimization strategy, the effect of the auxiliary policy gradient loss, and the necessity of correctness conditioning.

\vspace{-6pt}
\paragraph{Choice of Multi-View Optimization.}
We first conduct an ablation study evaluating the worst-case formulation (Eq. \ref{eq:worst}) for handling multi-view augmentations. Using the 8B model, we ablate this objective by replacing the worst-case penalty with a mean penalty across the augmented views. As detailed in Table \ref{tab:worst_case}, adopting this mean strategy incurs an average performance drop of 1.6\% on seen degradations and 1.8\% on unseen degradations compared to the worst-case formula. This demonstrates that averaging the invariance penalty is insufficient for securing robustness. Instead, by actively penalizing the hardest adversarial view during each optimization step, the worst-case formulation effectively forces the model to learn more robust reasoning capacity.

\begin{table*}[ht]
    \centering
    \caption{\textbf{Ablation on multi-view optimization strategy.} We evaluate the worst-case formulation by comparing it against averaging the invariance penalty across augmented views. Adopting the mean strategy incurs a performance drop for both seen and unseen degradations, demonstrating that actively penalizing the hardest adversarial view is better for robust reasoning.}
    
    \vspace{-3pt}
    \label{tab:worst_case}
    \resizebox{\textwidth}{!}{
    \begin{tabular}{l ccccccc c}
        \toprule
        \textbf{Method} & \textbf{MathVista} & \textbf{WeMath} &  \textbf{ChartQA} & \textbf{LogicVista} & \textbf{MMStar} & \textbf{VisPuzzles} & \textbf{RealWorldQA} & \textbf{Avg} \\
        \midrule
        \multicolumn{9}{c}{\textit{ Seen Degradations}} \\
        \midrule
        Mean Penalty & 72.1 & 75.1 & 49.0 & 53.5 & 64.6 & 40.9 & 65.1 & 60.0 \\
        Worst-Case Penalty & 73.3 & 77.3 & 49.1 & 57.5 & 66.3 & 41.7 & 66.0 & 61.6 \\

        \midrule
        \multicolumn{9}{c}{\textit{Unseen Degradations}} \\
        \midrule
        Mean Penalty & 63.2 & 67.2 & 44.0 & 47.5 & 60.1 & 37.8 & 62.0 & 54.5 \\
        Worst-Case Penalty & 64.8 & 70.4 & 44.1 & 50.8 & 62.1 & 38.0 & 63.6 & 56.3 \\
        \bottomrule
    \end{tabular}
    }
    \vspace{-8pt}
\end{table*}

\paragraph{Ablation on Auxiliary Policy Gradient.}

Next, we evaluate the contribution of the auxiliary policy gradient (PG) loss by removing it from the overall objective, with the 8B model. As shown in Table \ref{tab:pg_loss}, omitting this component reduces average accuracy by 1.6\% on seen degradations and 1.8\% on unseen degradations. This indicates that relying solely on the token-level invariance penalty is restrictive. While the invariance penalty successfully anchors the degraded output to the clean reference, it does not provide a sufficient learning signal to actively solve the reasoning task under visual occlusion. The auxiliary PG loss is therefore beneficial to provide a direct gradient signal that guides the policy toward correct reasoning steps despite the noise.

\begin{table*}[ht]
    \centering
    \caption{\textbf{Ablation on the auxiliary policy gradient loss.} We evaluate the framework's performance after removing the auxiliary PG component. Results indicate that integrating this loss improves reasoning accuracy across both seen and unseen degradations by providing a direct learning signal under visual occlusion.}
    \vspace{-3pt}
    \label{tab:pg_loss}
    \resizebox{\textwidth}{!}{
    \begin{tabular}{l ccccccc c}
        \toprule
        \textbf{Method} & \textbf{MathVista} & \textbf{WeMath} & \textbf{ChartQA} & \textbf{LogicVista} & \textbf{MMStar} & \textbf{VisPuzzles} & \textbf{RealWorldQA} & \textbf{Avg} \\
        \midrule
        \multicolumn{9}{c}{\textit{Seen Degradations}} \\
        \midrule
        w/o Auxiliary PG & 72.0 & 76.9 & 48.5 & 55.1 & 65.3 & 40.2 & 65.5 & 60.5 \\
        Full Approach & 73.3 & 77.3 & 49.1 & 57.5 & 66.3 & 41.7 & 66.0 & 61.6 \\
        \midrule
        \multicolumn{9}{c}{\textit{Unseen Degradations}} \\
        \midrule
        w/o Auxiliary PG & 63.9 & 69.5 & 43.6 & 49.4 & 60.8 & 37.9 & 62.6 & 55.4 \\
        Full Approach & 64.8 & 70.4 & 44.1 & 50.8 & 62.1 & 38.0 & 63.6 & 56.3 \\
        \bottomrule
    \end{tabular}
    }
\end{table*}

\vspace{-6pt}
\paragraph{Effect of Correctness Conditioning.}

\begin{wraptable}{r}{0.4\textwidth}
\vspace{-12pt}
    \centering
    \caption{Sensitivity analysis of the auxiliary coefficient $\alpha$ on seen and unseen degradations. The best performance is achieved at $\alpha = 0.10$.}
    \vspace{-3pt}
    \label{tab:alpha}
    \resizebox{\linewidth}{!}{
    \begin{tabular}{lcc}
    \toprule
    \small
    \textbf{Auxiliary Coefficient} & \textbf{Seen} & \textbf{Unseen} \\
    \midrule
    $\alpha=0.05$ & 60.5 & 55.2 \\
    $\alpha=0.10$ & 61.6 & 56.3 \\
    $\alpha=0.15$ & 60.0 & 54.8 \\
   \bottomrule
\end{tabular}
}
\vspace{-12pt}
\end{wraptable}

Finally, we investigate the role of correctness conditioning within the token-level KL penalty using the 8B model. As detailed in Table \ref{tab:cc}, enforcing the invariance penalty unconditionally, forcing the degraded reasoning trajectory to match the clean trajectory regardless of whether the clean rationale is correct, causes an average performance drop of 2.2\% on both seen and unseen degradations. By conditioning the penalty on the objective correctness of the clean rollout, our approach ensures that the policy learns to protect valid reasoning paths, effectively preventing the propagation of erroneous logic during the optimization process.

\begin{table*}[ht]
    \centering
    \caption{\textbf{Ablation on correctness conditioning.} We evaluate the necessity of conditioning the invariance penalty on the objective correctness of the clean trajectory. Results demonstrate that strictly applying the penalty to valid reasoning paths prevents the propagation of erroneous logic and improves overall performance.}
    \vspace{-3pt}
    \label{tab:cc}
    \resizebox{\textwidth}{!}{
    \begin{tabular}{l ccccccc c}
        \toprule
        \textbf{Method} & \textbf{MathVista} & \textbf{WeMath} &  \textbf{ChartQA} & \textbf{LogicVista} & \textbf{MMStar} & \textbf{VisPuzzles} & \textbf{RealWorld} & \textbf{Avg} \\
        \midrule
        \multicolumn{9}{c}{\textit{Seen Degradations}} \\
        \midrule
        Unconditional Penalty & 69.8 & 74.7 & 47.9 & 52.6 & 65.0 & 40.0 & 65.5 & 59.4 \\
        Correctness-Conditioned & 73.3 & 77.3 & 49.1 & 57.5 & 66.3 & 41.7 & 66.0 & 61.6 \\
        \midrule
        \multicolumn{9}{c}{\textit{ Unseen Degradations}} \\
        \midrule
        Unconditional Penalty & 61.8 & 67.0  & 43.2 & 48.2 & 59.8 & 36.8 & 62.1 & 54.1 \\
        Correctness-Conditioned & 64.8 & 70.4 & 44.1 & 50.8 & 62.1 & 38.0 & 63.6 & 56.3 \\
        \bottomrule
    \end{tabular}
    }
    \vspace{-10pt}
\end{table*}

\vspace{-6pt}
\subsection{Sensitivity Analysis} \label{subsec:sensitivity_analysis}
\vspace{-6pt}

Wo conduct a series of sensitivity analysis to the key hyperparameters with the 8B model, including the auxiliary policy gradient coefficient $\alpha$, the number of augmented views $K$, and the invariance penalty weight $\beta$. 

\vspace{-6pt}
\paragraph{Auxiliary Coefficient.} 

We evaluate the framework's sensitivity to the auxiliary policy gradient loss by varying the coefficient $\alpha \in \{0.05, 0.10, 0.15\}$. As presented in Table \ref{tab:alpha}, the model achieves best performance at $\alpha=0.10$, yielding 61.6\% and 56.3\% accuracy on seen and unseen degradations, respectively. Decreasing the coefficient to $\alpha=0.05$ provides insufficient auxiliary guidance, resulting in a performance drop. Conversely, increasing $\alpha$ to 0.15 also leads to a reduction, as the excessively weighted auxiliary loss begins to over-regularize and interfere with the primary optimization objective.

\vspace{-6pt}
\paragraph{Number of Augmented Views.} 

\begin{wraptable}{r}{0.4\textwidth}
\vspace{-12pt}
    \centering
    \caption{Sensitivity analysis of the number of augmented views $K$ on seen and unseen degradations. We select $K = 3$ as the default setting.}
    \vspace{-5pt}
    \label{tab:k}
    \resizebox{\linewidth}{!}{
    \begin{tabular}{lcc}
    \toprule
    \small
    \textbf{Number of Augmented Views} & \textbf{Seen} & \textbf{Unseen} \\
    \midrule
    $K=1$ & 59.5 & 54.4 \\
    $K=2$ & 60.7 & 55.2 \\
    $K=3$ & 61.6 & 56.3 \\
    $K=4$ & 61.3 & 56.0 \\
   \bottomrule
\end{tabular}
}
\vspace{-10pt}
\end{wraptable}

The parameter K dictates the diversity of perturbations evaluated during the optimization step. As shown in Table \ref{tab:k}, increasing K from 1 to 3 yields steady improvements in both seen and unseen robustness, as the policy is penalized against a broader distribution of visual noise. However, increasing K beyond 3 (e.g., $K=4$) provides slight performance degradation. Therefore, we select $K=3$ as the default setting to maintain a computationally efficient training pipeline without sacrificing robust generalization.

\paragraph{Invariance Penalty Weight.} 

\begin{wraptable}{r}{0.4\textwidth}
\vspace{-12pt}
    \centering
    \caption{Sensitivity analysis of the penalty weight $\beta$ on seen and unseen degradations. The best performance is achieved at $\beta = 0.10$.}
    \vspace{-5pt}
    \label{tab:alpha}
    \resizebox{\linewidth}{!}{
    \begin{tabular}{lcc}
    \toprule
    \small
    \textbf{Auxiliary Coefficient} & \textbf{Seen} & \textbf{Unseen} \\
    \midrule
    $\beta=0.05$ & 59.4 & 55.3 \\
    $\beta=0.10$ & 61.3 & 56.3 \\
    $\beta=0.15$ & 56.8 & 54.6 \\
   \bottomrule
\end{tabular}
}
\vspace{-10pt}
\end{wraptable}

We investigate the trade-off between baseline reasoning capacity and visual robustness by varying $\beta \in \{0.05, 0.10, 0.15\}$. A high penalty weight (e.g., $\beta = 0.15$) overly constrains the policy, forcing it to prioritize structural matching over exploratory problem-solving, which leads to a performance drop. Conversely, a small weight fails to enforce sufficient noise resilience. We find that $\beta = 0.10$ establishes an optimal balance, maximizing robustness without degrading foundational reasoning capabilities.

\vspace{-6pt}
\section{Conclusions} \label{sec:con}
\vspace{-6pt}
Reinforcement Learning has significantly advanced the reasoning capabilities of MLLMs, yet these models remain brittle when faced with real-world visual degradations. Standard robustness techniques struggle with the architectural constraints of large-scale, critic-free RL and the risk of reward poisoning, where perceptual occlusions lead to policy collapse. To address these challenges, we propose a novel RL fine-tuning framework that integrates adversarial visual robustness directly into the reasoning pipeline. Our approach employs a dual-forward-pass strategy, utilizing teacher-forcing to evaluate corrupted views against trajectories generated from clean images. We introduce a token-level KL divergence penalty on worst-case visual augmentations to ensure distributional consistency, complemented by an auxiliary policy gradient loss that preserves reward signals under degradation. Our method enables MLLMs to internalize robust logic, maintaining reasoning stability across diverse visual corruptions without sacrificing performance on clean data. Please refer to a discussion on future work in Appendix \ref{app:future}. 

\bibliography{ref}
\bibliographystyle{plainnat}

\newpage
\appendix
\section{Appendix}

\subsection{Implementation Details} \label{app:imp}

We train all models on the $\text{MMRL30k}$ dataset \citep{zhu2025shuffle}, which contains around 30K samples. The models are trained to generate responses in a structured format, where the reasoning process is enclosed within \(\texttt{<thinking></thinking>}\) tags and the final answer is presented in \texttt{\textbackslash boxed\{\}}. The training is performed for 120 steps with a learning rate of $1e-6$ and a weight decay of $0.01$. We adopt a global batch size of $128$, a rollout batch size of $256$, and generate $8$ rollouts per input with a rollout temperature $1.0$. The implementation builds on the EasyR1 framework \citep{zheng2025easyr1}.

\subsection{Degradation Details and Severity Levels} \label{app:deg}
\begin{figure}[ht]
    \centering
    \includegraphics[width=\linewidth]{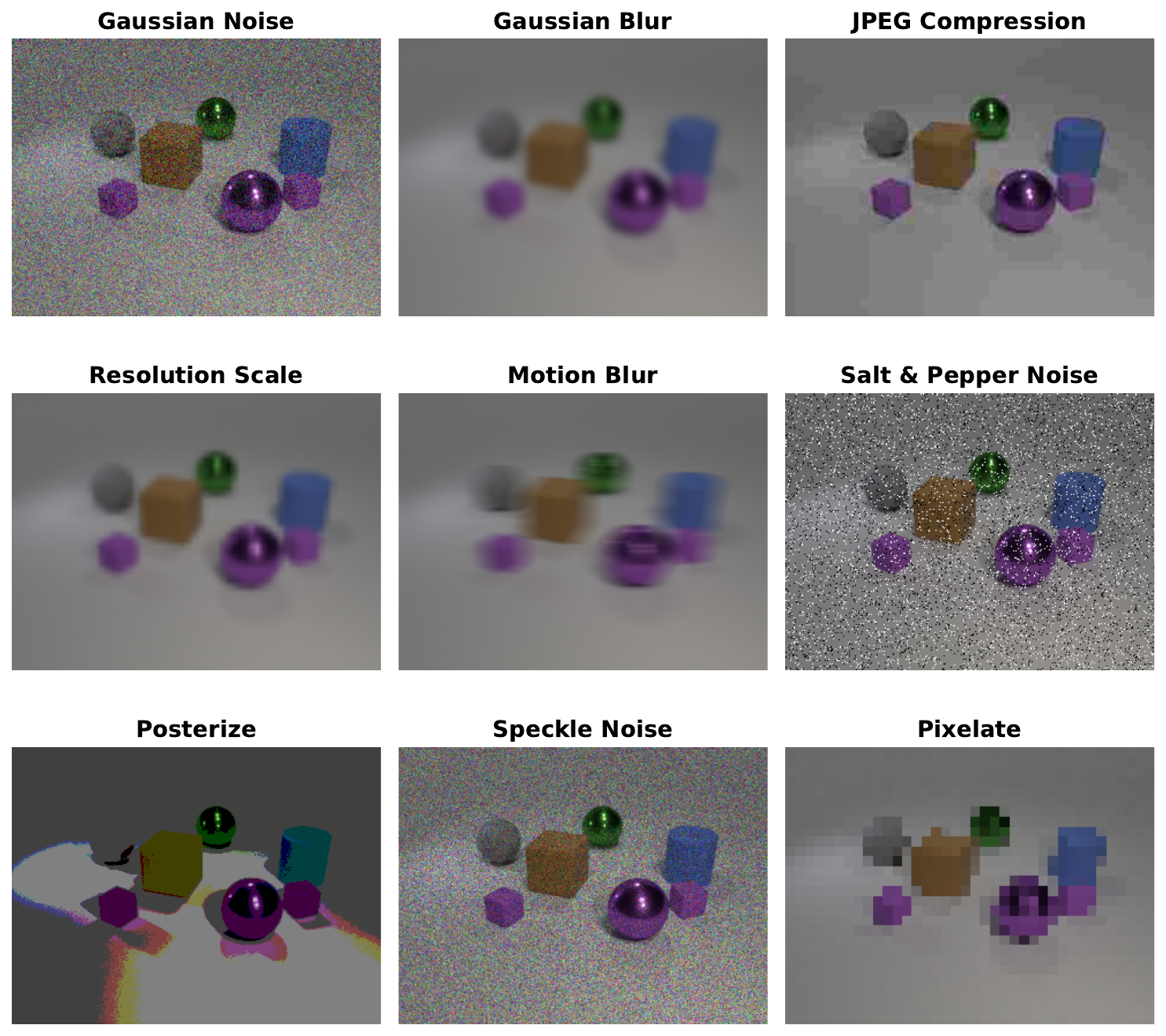}
    \caption{\textbf{Qualitative examples of visual degradations.} The perturbations are categorized into seen degradations (e.g., Gaussian noise, Gaussian blur, JPEG compression, and Resolution downscaling), which simulate common transmission artifacts encountered during training, and unseen degradations (e.g., motion blur, salt-and-pepper noise, speckle noise, posterization, and pixelation), which test out-of-distribution (OOD) generalization against held-out noise structures.}
    \label{fig:deg}
\end{figure}

To rigorously evaluate the robustness of our approach, we apply a diverse set of visual corruptions. During training, parameters are sampled continuously according to corresponding distributions. During evaluation, we utilize three severity levels (Level 1 to Level 3) for benchmarking, following the ImageNet-C framework \citep{hendrycks2019benchmarking}. Crucially, Level 3 is designed to evaluate OOD magnitude generalization. For every degradation type, Level 3 applies a severity that strictly exceeds the bounds of the parameter distribution encountered by the model during training. The parameter configurations for the degradations are detailed below in Table \ref{tab:degradation_params} and qualitative examples are provided in Figure \ref{fig:deg}.

\begin{table}[h]
  \centering
  \caption{Parameter configurations for visual degradations during training and evaluation. For JPEG Quality, Resolution Scale, Posterization, and Pixelation, a lower value indicates higher severity. For the seen pool, Level 3 consistently falls outside the bounds of the training distribution (magnitude OOD). The unseen pool is strictly OOD across all levels.}
  \label{tab:degradation_params}
  \resizebox{\columnwidth}{!}{%
  \begin{tabular}{ll cccc}
  \toprule
  \textbf{Degradation Type} & \textbf{Parameter} & \textbf{Training Distribution} & \textbf{Eval Level 1} & \textbf{Eval Level 2} &
  \textbf{Eval Level 3} \\ \midrule
  \multicolumn{6}{c}{\textit{Seen Degradations}} \\ \midrule
  Gaussian Noise & Std. Dev. ($\sigma$) & $0.05$ & $0.03$ & $0.06$ & $\mathbf{0.12}$ \\
  Gaussian Blur & Radius ($r$) & $\mathcal{U}(0.5, 2.0)$ & $1.0$ & $2.0$ & $\mathbf{3.5}$ \\
  JPEG Compression & Quality ($q$) $\downarrow$ & $\mathcal{U}\{30, 85\}$ & $65$ & $40$ & $\mathbf{15}$ \\
  Resolution Scale & Scale Factor ($f$) $\downarrow$ & $\mathcal{U}(0.3, 0.7)$ & $0.6$ & $0.4$ & $\mathbf{0.2}$ \\ \midrule
  \multicolumn{6}{c}{\textit{Unseen Degradations}} \\ \midrule
  Motion Blur & Kernel Size ($k$) & (Held Out) & $5$ & $15$ & $25$ \\
  Salt \& Pepper Noise & Probability ($p$) & (Held Out) & $0.02$ & $0.05$ & $0.10$ \\
  Speckle Noise & Std. Dev. ($\sigma$) & (Held Out) & $0.06$ & $0.12$ & $0.25$ \\
  Posterization & Bit Depth ($b$) $\downarrow$ & (Held Out) & $6$ & $4$ & $2$ \\
  Pixelation & Scale Factor ($f$) $\downarrow$ & (Held Out) & $0.5$ & $0.25$ & $0.12$ \\
  \bottomrule
  \end{tabular}%
  }
\end{table}


\subsection{Experiments} \label{app:exp}
We present the detailed evaluation results of the 8B model across all seen and unseen degradation types in Tables \ref{tab:detailed_8b_base}, \ref{tab:detailed_8b_grpo}, and \ref{tab:detailed_8b_ours} for the base model, the GRPO baseline, and our approach, respectively.

\begin{table*}[ht]
    \centering
    \caption{Detailed performance breakdown of the Base 8B model across specific visual degradations.}
    \label{tab:detailed_8b_base}
    \resizebox{\textwidth}{!}{
    \begin{tabular}{l ccccccc}
        \toprule
        \textbf{Perturbation} & \textbf{MathVista} & \textbf{WeMath} & \textbf{ChartQA} & \textbf{LogicVista} & \textbf{MMStar} & \textbf{VisPuzzles} & \textbf{RealWorldQA} \\
        \midrule
        Clean & 76.6 & 69.6 & 79.4 & 60.7 & 68.7 & 43.5 & 69.4 \\
        \midrule
        \multicolumn{8}{c}{\textit{Seen Degradations}} \\
        \midrule
        Gaussian Blur & 67.0 & 68.7 & 14.6 & 49.8 & 60.3 & 39.2 & 64.1 \\
        Gaussian Noise & 73.5 & 69.6 & 77.5 & 58.3 & 68.0 & 41.4 & 67.5 \\
        JPEG Compression & 75.4 & 69.9 & 77.2 & 56.7 & 67.0 & 42.0 & 67.5 \\
        Resolution Scale & 67.0 & 69.3 & 18.2 & 53.8 & 63.9 & 37.9 & 63.8 \\
        Avg & 70.7 & 69.4 & 46.9 & 54.6 & 64.8 & 40.1 & 65.7 \\ 
        \midrule
        \multicolumn{8}{c}{\textit{Unseen Degradations}} \\
        \midrule
        Motion Blur & 55.8 & 55.4 & 6.4 & 42.4 & 54.1 & 36.2 & 60.1 \\
        Pixelate & 45.2 & 39.5 & 5.4 & 35.3 & 49.6 & 27.7 & 54.4 \\
        Posterize & 71.0 & 69.2 & 77.8 & 58.0 & 64.2 & 39.7 & 65.2 \\
        Salt \& Pepper Noise & 68.2 & 69.1 & 45.4 & 55.6 & 61.9 & 40.9 & 60.9 \\
        Speckle Noise & 75.2 & 68.6 & 75.6 & 58.5 & 67.7 & 41.8 & 66.4 \\
        Avg & 63.1 & 60.4 & 42.1 & 50.0 & 59.5 & 37.3 & 61.4 \\
        \bottomrule
    \end{tabular}
    }
\end{table*}


\begin{table*}[ht]
    \centering
    \caption{Detailed performance breakdown of the 8B model across specific visual degradations using the GRPO baseline.}
    \label{tab:detailed_8b_grpo}
    \resizebox{\textwidth}{!}{
    \begin{tabular}{l ccccccc}
        \toprule
        \textbf{Perturbation} & \textbf{MathVista} & \textbf{WeMath} & \textbf{ChartQA} & \textbf{LogicVista} & \textbf{MMStar} & \textbf{VisPuzzles} & \textbf{RealWorldQA} \\
        \midrule
        Clean & 78.4 & 77.6 & 81.5 & 60.8 & 70.1 & 43.5 & 70.6 \\
        \midrule
        \multicolumn{8}{c}{\textit{Seen Degradations}} \\
        \midrule
        Gaussian Blur & 67.5 & 74.4 & 15.8 & 47.8 & 59.5 & 38.0 & 63.2 \\
        Gaussian Noise & 74.0 & 75.2 & 78.6 & 56.4 & 67.2 & 40.2 & 66.5 \\
        JPEG Compression & 75.8 & 75.5 & 78.3 & 54.8 & 66.2 & 40.7 & 66.6 \\
        Resolution Scale & 67.5 & 74.9 & 19.3 & 51.8 & 63.1 & 36.7 & 62.9 \\
        Avg & 71.2 & 75.0 & 48.0 & 52.7 & 64.0 & 38.9 & 64.8 \\
        \midrule
        \multicolumn{8}{c}{\textit{Unseen Degradations}} \\
        \midrule
        Motion Blur & 55.7 & 61.8 & 7.3 & 39.5 & 54.4 & 35.7 & 60.1 \\
        Pixelate & 45.2 & 46.1 & 6.3 & 32.6 & 49.9 & 27.4 & 54.4 \\
        Posterize & 70.9 & 75.6 & 78.6 & 55.1 & 64.5 & 39.2 & 65.2 \\
        Salt \& Pepper Noise & 68.1 & 75.5 & 46.3 & 52.7 & 62.2 & 40.4 & 60.9 \\
        Speckle Noise & 75.1 & 75.0 & 76.5 & 55.6 & 68.0 & 41.3 & 66.4 \\
        Avg & 63.0 & 66.8 & 43.0 & 47.1 & 59.8 & 36.8 & 61.4 \\
        \bottomrule
    \end{tabular}
    }
\end{table*}

\begin{table*}[ht]
    \centering
    \caption{Detailed performance breakdown of the 8B model across specific visual degradations using our approach.}
    \label{tab:detailed_8b_ours}
    \resizebox{\textwidth}{!}{
    \begin{tabular}{l ccccccc}
        \toprule
        \textbf{Perturbation} & \textbf{MathVista} & \textbf{WeMath} & \textbf{ChartQA} & \textbf{LogicVista} & \textbf{MMStar} & \textbf{VisPuzzles} & \textbf{RealWorldQA} \\
        \midrule
        Clean & 78.5 & 77.9 & 80.8 & 62.1 & 69.5 & 42.5 & 69.9 \\
        \midrule
        \multicolumn{8}{c}{\textit{Seen Degradations}} \\
        \midrule
        Gaussian Blur & 69.6 & 76.7 & 16.9 & 52.6 & 61.8 & 40.8 & 64.4 \\
        Gaussian Noise & 76.1 & 77.5 & 79.7 & 61.2 & 69.5 & 43.0 & 67.7 \\
        JPEG Compression & 77.9 & 77.8 & 79.4 & 59.6 & 68.5 & 43.5 & 67.8 \\
        Resolution Scale & 69.6 & 77.2 & 20.4 & 56.6 & 65.4 & 39.5 & 64.1 \\
        Avg & 73.3 & 77.3 & 49.1 & 57.5 & 66.3 & 41.7 & 66.0 \\
        \midrule
        \multicolumn{8}{c}{\textit{Unseen Degradations}} \\
        \midrule
        Motion Blur & 57.5 & 65.4 & 8.4 & 43.2 & 56.7 & 36.9 & 62.3 \\
        Pixelate & 47.0 & 49.7 & 7.4 & 36.3 & 52.2 & 28.6 & 56.6 \\
        Posterize & 72.7 & 79.2 & 79.7 & 58.8 & 66.8 & 40.4 & 67.4 \\
        Salt \& Pepper Noise & 69.9 & 79.1 & 47.4 & 56.4 & 64.5 & 41.6 & 63.1 \\
        Speckle Noise & 76.9 & 78.6 & 77.6 & 59.3 & 70.3 & 42.5 & 68.6 \\
        Avg & 64.8 & 70.4 & 44.1 & 50.8 & 62.1 & 38.0 & 63.6 \\
        \bottomrule
    \end{tabular}
    }
\end{table*}


\subsection{Discussions and Future Work} \label{app:future}

While our proposed approach establishes a robust foundation for multimodal reasoning against degradations, it also opens several promising avenues for future research. A natural progression is to extend this worst-case multi-view optimization paradigm to temporal modalities, such as video-based reasoning. Furthermore, future work could investigate adaptive mechanisms to dynamically weight both the auxiliary policy gradient objective and the invariance penalty based on the inferred severity of the visual degradation, thereby allocating stronger defensive penalties specifically to highly adversarial inputs.

\clearpage

\end{document}